\documentclass[wcp]{jmlr}

\usepackage{longtable}

\usepackage{booktabs}

\makeatletter
\let\Ginclude@graphics\@org@Ginclude@graphics 

\renewcommand*{\thanks}[1]{%
  \footnotemark
  \protected@xdef\@thanks{\@thanks
    \protect\footnotetext[\arabic{footnote}]{#1}}%
}

\makeatother

\jmlrvolume{189}
\jmlryear{2022}
\jmlrworkshop{ACML 2022}

\title[Semantic Cross Attention for Few-shot Learning]{Semantic Cross Attention for Few-shot Learning}

 \author{\Name{Bin Xiao}\thanks{Bin Xiao finished this work when he was at National Chiao Tung University} \Email{bxiao103@uottawa.ca}\\
  \addr School of Electrical Engineering and Computer Science, University of Ottawa, 535 Legget Drive, Kanata, K2K 3B8, Ottawa, Canada
  \AND
  \Name{Chien-Liang Liu}\thanks{Corresponding author} \Email{clliu@nycu.edu.tw}\\
  \addr Department of Industrial Engineering and Management, National Yang Ming Chiao Tung University, 1001 University Road, Hsinchu 30010, Taiwan, ROC
  \AND
  \Name{Wen-Hoar Hsaio} \Email{bass228@nanya.edu.tw}\\
  \addr Department of Computer Science and Engineering,  Nanya Institute of Technology, Taoyuan 32091, Taiwan, ROC
 }
 
\editors{Emtiyaz Khan and Mehmet G\"{o}nen}

\begin{document}

\maketitle

\begin{abstract}
Few-shot learning (FSL) has attracted considerable attention recently. Among existing approaches, the metric-based method aims to train an embedding network that can make similar samples close while dissimilar samples as far as possible and achieves promising results. FSL is characterized by using only a few images to train a model that can generalize to novel classes in image classification problems, but this setting makes it difficult to learn the visual features that can identify the images' appearance variations. The model training is likely to move in the wrong direction, as the images in an identical semantic class may have dissimilar appearances, whereas the images in different semantic classes may share a similar appearance. We argue that FSL can benefit from additional semantic features to learn discriminative feature representations. Thus, this study proposes a multi-task learning approach to view semantic features of label text as an auxiliary task to help boost the performance of the FSL task. Our proposed model uses word-embedding representations as semantic features to help train the embedding network and a semantic cross-attention module to bridge the semantic features into the typical visual modal. The proposed approach is simple, but produces excellent results. We apply our proposed approach to two previous metric-based FSL methods, all of which can substantially improve performance. The source code for our model is accessible from github~\footnote{\url{https://github.com/uobinxiao/semantic_cross_attention_fsl}}.
\end{abstract}

\section{Introduction}
\begin{figure}[htp]
\begin{center}
  \includegraphics[width=\textwidth]{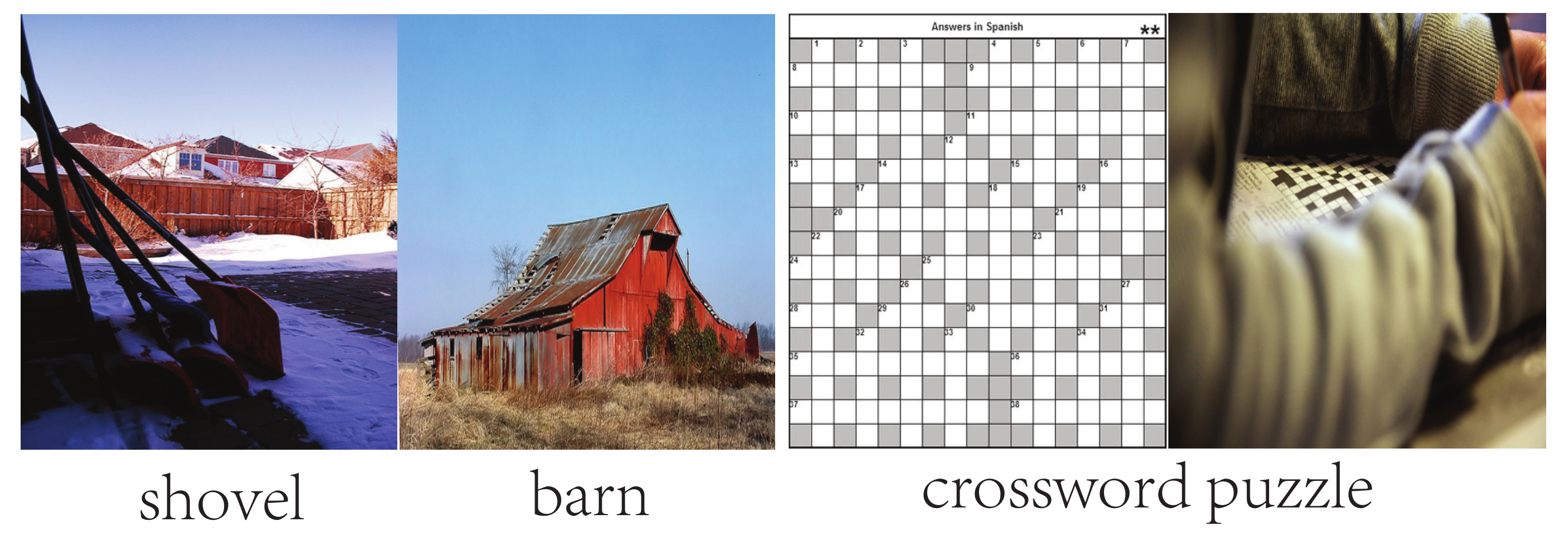}
  \caption{Examples of a similar visual appearance of images in different classes (left) and a dissimilar visual appearance of images in an identical class (right). The images and their labels are from the tiered-ImageNet dataset.}
  \label{fig:visual_features}
\end{center}
\end{figure}

Few-shot learning (FSL) algorithms have been widely studied in recent years, and most of the work has focused on the few-shot image classification problem. For the few-shot image classification problem, it is a natural choice to use convolution neural networks (CNN)~\citep{Lecun1998,krizhevsky2012imagenet} to extract visual features from the images. However, simply using visual features can lead to the following problems when training samples involved in the training process are limited. First, images from different classes may share a similar visual appearance. For example, Figure~\ref{fig:visual_features} shows that the shovel and the barn share a similar visual appearance, but their labels appear different, which means that their semantic meanings are different. Second, images from an identical class may have dissimilar visual appearances. The images of the crossword puzzle as shown in Figure~\ref{fig:visual_features} present this problem. Although these two crossword puzzle images have identical semantic meanings, their visual appearances are different. We call these two problems a visual-semantic mismatch, in which ``visual'' means the visual features learned from deep learning models and ``semantic'' denotes their semantic labels. In particular, the images and labels in Figure~\ref{fig:visual_features} are all from the tiered-ImageNet dataset~\citep{ren2018meta}, which is a widely used dataset in the few-shot image classification problem. It is worth mentioning that a similar concept called Shortcut Learning~\citep{geirhos2020shortcut, luo2021rectifying} has been discussed in other studies, whose main motivation is to guide the model to learn intended features rather than shortcut features. In the context of few-shot learning, \citet{luo2021rectifying} considered the impact of the image background that may contain the shortcut information. However, the motivation of this study is to handle the visual-semantic mismatch as mentioned above, so our proposed method is different from studies considering short-cut learning.

The aforementioned visual-semantic mismatch problem is not present in an image classification model trained with a large-scale dataset and augmentation techniques. However, in the FSL setting, the number of images in each class is very limited, so the visual features learned by the model are difficult to reflect the appearance variations of the images. The classical metric-based method for FSL relies on an embedding network to transform the input images into feature embeddings that can make similar samples close while dissimilar samples as far as possible. Many previous works~\citep{chen2019closerfewshot,xiao2020proxy,tian2020rethinking} have empirically shown that the embedding network is crucial for performance. Therefore, this study proposes a multi-task learning model to focus on learning feature embeddings that can quickly adapt to novel classes by leveraging an auxiliary task. The auxiliary task incorporates semantic features into the training process, and the goal is to further refine the embedding network. In addition, we propose a semantic cross-attention module (CAM) to guide the visual embedding network to focus on the correct semantic areas. More specifically, we use a word-embedding model to transform text labels into soft labels as the target of the auxiliary task to inject semantic features into the main task. It is worth mentioning that the proposed CAM module firstly flattens the visual features into a patch sequence and then maps the patch sequence to the key and value vectors. Still, the query vectors are generated by the semantic auxiliary task. Therefore, the proposed CAM module is different from the self-attention mechanism. Once the three types of vectors are available, the dot product of the query vectors and the key vectors is used to generate the weight matrices, and the final output is based on the weight matrices and the value vectors. This way, semantic features can be injected into visual features, guiding the embedding network to focus on the correct areas. 

The contributions of this work are three-fold. First, we inspect the problem of FSL and present the visual-semantic mismatch problem that is present in the FSL setting. It is difficult to directly tackle this problem by only using limited images, which is why we propose to incorporate semantic features into the model to help alleviate the problem of only limited images available for each episode. Second, we simultaneously consider the semantic meaning of text labels and visual features to propose a multi-task learning model that can align visual features and their semantic meanings. Finally, we apply our proposed method to two recent metric-based methods. The experimental results indicate that these methods can benefit from our proposed method and yield substantial improvements in the experiments.

\section{Related work}
\subsection{Few-shot learning}

Few-shot learning methods can be roughly classified into three groups: metric-based methods, optimization-based methods, and hallucination-based methods~\citep{chen2019closerfewshot}. Our proposed method uses metric-based methods as the base model, so this section focuses on metric-based approaches. The metric-based FSL method usually comprises three key components, including the embedding network, the class representatives, and the distance metric~\citep{xiao2020proxy}. For example, ProtoNet~\citep{snell2017prototypical} is a classical metric-based method and comprises an embedding network to transform each input image into a visual embedding vector. In addition, it uses the prototype of each class's feature embeddings as the class representative and uses Euclidean distance as the distance metric. \citet{ren2018meta} extended ProtoNet in a semi-supervised manner and discussed two semi-supervised few-shot classification settings. \citet{Liu2020PrototypeRF} proposed a transductive ProtoNet by introducing a bias diminishing module, which tries to refine the feature embeddings by introducing an extra module. They pointed out that the intra-class bias and the cross-class bias are two key points that can influence the representativeness of class prototypes and proposed a bias diminishing module by using pseudo-labeling and feature shifting. However, their proposed method works in a transductive setting, which differs from most FSL methods. Tian et al. proposed a method based on knowledge distillation to improve feature embeddings~\citep{tian2020rethinking}. Instead of using a typical teacher model, they proposed using the trained model at different epochs as a teacher model, also called self-distillation. Many works~\citep{chen2019closerfewshot,tian2020rethinking} have also discussed the influence of a deeper embedding network and concluded that the model could generally benefit from a deeper embedding network to achieve better performance to some extent. 

RelationNet~\citep{sung2018learning} improves the metric-based method by using a trainable distance metric, which comprises convolution layers and fully connected layers. Similarly, ProxyNet~\citep{xiao2020proxy} also defines a trainable metric using a 3D convolution layer and uses trainable class representatives. \citet{simon2020adaptive} proposed to use dynamic classifiers that are based on a deep subspace network to extract a discriminated subspace from the feature embeddings. The goal is to further improve the distance metric for typical metric-based methods.

Several studies have considered using the features of the text modality to deal with the FSL problem. \citet{xing2019adaptive} also pointed out the drawbacks of using only visual features to deal with FSL and proposed to use the semantic features of the text modality. They proposed an adaptive modality mixture mechanism (AM3) that can combine the features of visual modality and text modality adaptively and selectively. Although AM3 also considers text and visual modalities, the purpose and approach are different from our proposed approach. More specifically, AM3 uses a weighted sum of visual and semantic features as prototypes, and the weights are determined by an adaptive mixing network. The semantic features generated from class labels are required input for AM3 in both meta-training and meta-test stages, since the model relies on the semantic features to calculate the prototypes. In contrast, our proposed method is based on a multi-task architecture and focuses on learning visual-semantic features that can quickly adapt to novel classes by leveraging an auxiliary task. Besides, we propose a CAM module to inject semantic features into visual embeddings. It is worth mentioning that label information is not required in our meta-testing stage, as the main task is a few-shot classification. The study conducted by \citet{pahde2021multimodal} is also a multi-modality method and uses a generative adversarial network (GAN) framework to combine the features of the text modality with the visual features. Moreover, this model takes images and text descriptions as inputs, and the model comprises the semantic features of the input text description. Notably, the text description is an additional knowledge source, which differs from the label that is available in the meta-training set.

\subsection{Multi-task learning}

Multi-task learning is a learning paradigm that optimizes several tasks simultaneously, and the goal is to use auxiliary tasks to improve the performance of the main task. The role of auxiliary tasks is similar to that of a regularization term that can place additional constraints on the model. Additionally, the main task can benefit from the parameter sharing mechanism to improve the performance of the model. Many approaches have been proposed by designing different parameter sharing mechanisms. One of the most widely used approaches is hard parameter sharing~\citep{Caruana1993MultitaskLA,ruder2017overview}. For deep learning models, hard parameter sharing can be realized by using sharing layers to learn the feature representations that are common for all tasks and by keeping task-specific layers to learn the representations that are specific for all tasks. This approach is simple and easy to implement, which is why many studies have used the hard parameter sharing approach to develop methods and achieved promising performance~\citep{zhang2014facial,dai2016instance}. In contrast, soft parameter sharing~\citep{ruder2017overview} allows each task to have its own model with its own parameters and uses a regularization technique to make the parameters of the tasks similar.

Besides, multi-task learning can be viewed as a transfer learning approach that transfers knowledge between different tasks. Note that multi-task learning cannot guarantee that the main task can benefit from the auxiliary tasks to improve performance, as negative transfer may occur, which can degrade the performance of the main task. To alleviate negative transfer, \citet{lee2016asymmetric} proposed a method called asymmetric multi-task learning (AMTL) by considering the task relatedness of the task and the loss of each task to measure the reliability of each task. In addition to task loss, other metrics, such as uncertainty~\citep{Nguyen2020ClinicalRP,Kendall2017WhatUD}, can also be used to measure task reliability.

\section{Proposed Method}
This study follows conventional meta-learning settings to train the proposed method, so the model training is processed episodically. In each episode, the input comprises a sequence of meta-tasks, each of which comprises a support set and a query set. The images for these two sets are non-overlapped and are randomly drawn from the meta-training set. In particular, we use meta-task to represent the tasks in episode-based meta-training to distinguish with the \emph{task} that is used by multi-task learning in this section.

For a $K$-way-$N$-shot problem, the support set $S = \{(\mathbf{x}_i, y_i)\}_{i=1}^{K \cdot N}$ comprises $K \cdot N$ images that are randomly selected from $K$ classes, each of which has $N$ images. On the other hand, the query set $Q=\{(\mathbf{x}_j, y_i)\}_{j=1}^{M}$ comprises $M$ images that are randomly sampled from the corresponding $K$ classes and differ from the images of the support set. These sampled meta-tasks can be not only the inputs of the main task, but also the inputs of the auxiliary task in our multi-task learning model.

The proposed method can be applied to different metric-based methods. To explain how our proposed model works, we use ProtoNet~\citep{snell2017prototypical} as an example to further illustrate the proposed method, since ProtoNet is a classical metric-based method for FSL. In the experiments, we use different metric-based methods as base methods to show the performance improvement brought about by our proposed method.

\begin{figure*}[htp]
\begin{center}
  \includegraphics[width=1.0\textwidth]{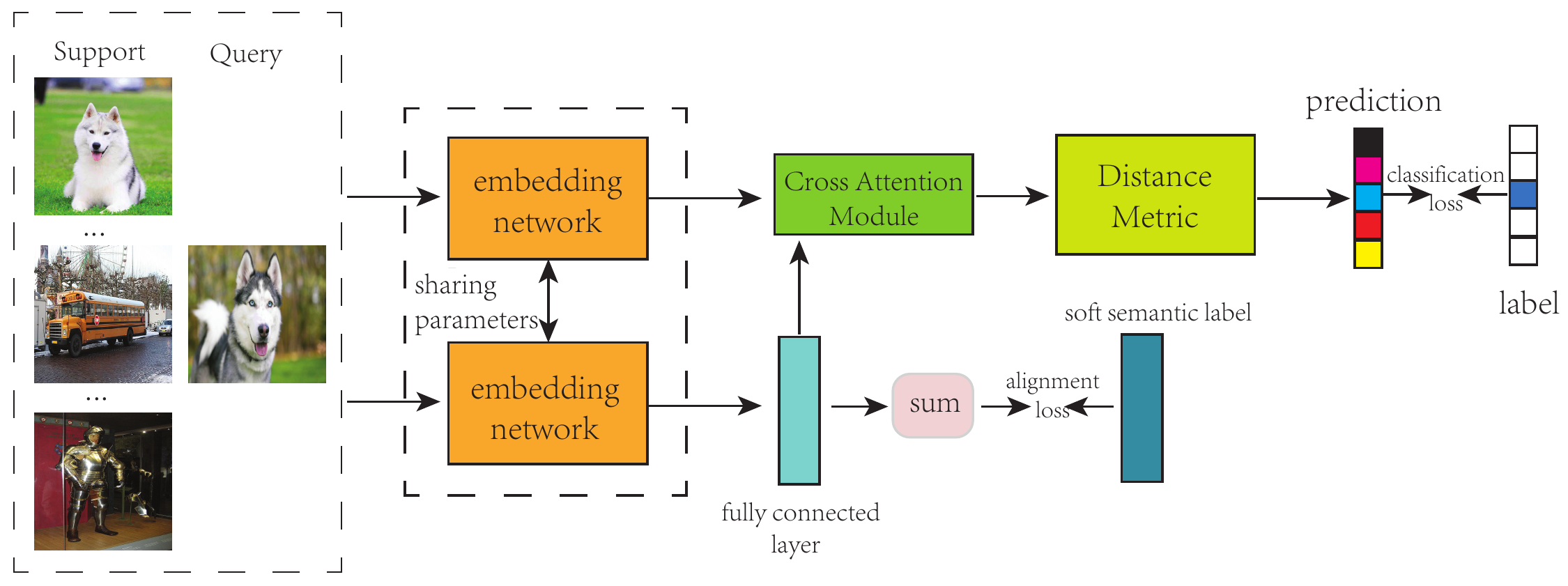}
  \caption{The overall learning architecture of the proposed multi-task model.}
  \label{fig:overal_architecture}
\end{center}  
\end{figure*}

The embedding network plays a crucial role in metric-based methods~\citep{chen2019closerfewshot,xiao2020proxy,tian2020rethinking}, which is why this study aims to incorporate semantic features into the model to help train an embedding network that can adapt to novel classes. Unlike other metric-based methods that only use images to train the embedding network, this study proposes using semantic features to enhance the training of the embedding network, in which the proposed method comprises text and image modalities, and the text modality can provide semantic information. To incorporate two modalities into the model, we propose to use a multi-task learning architecture with two tasks to handle visual features and the semantic information of text labels, respectively.

Figure~\ref{fig:overal_architecture} shows the overall architecture of the proposed method, which contains two tasks: the main task for the visual modality and the auxiliary task for the text modality. The former deals with few-shot classification, while the latter aims to incorporate semantic features into the model. We use the hard parameter sharing method to design the model so that the main task and the auxiliary task share an identical embedding network $\mathcal{E}_\theta$. For the main task, an input image $x$ is first transformed into feature maps using the embedding network $\mathcal{E}_\theta$; then, the feature maps are fed into the CAM denoted by $Cross_\phi$ to obtain the final embedding of the features of each image. The entire process can be represented by Equation~\eqref{eq:embeddng_network} and Equation~\eqref{eq:cross_attention}, in which $ch_{in}, h_{in}, w_{in}$ are the number of channels, height and width of the input image, respectively, $l_{inter}$ is the depth of the feature map generated by $\mathcal{E}_\theta$, and $l_{out}, h_{out}, w_{out}$ are the depth, height and width of the final embedding generated by $Cross_\phi$. It is worth mentioning that $Cross_\phi$ also takes an input $e_{aux}$ from the auxiliary task.

\begin{equation}
\label{eq:embeddng_network}
e_{main} = \mathcal{E}_\theta(x), x \in \mathbb{R}^{ch_{in} \times h_{in} \times w_{in}}, e \in \mathbb{R}^{l_{inter} \times h_{out} \times w_{out}}
\end{equation}

\begin{equation}
\label{eq:cross_attention}
e_{out} = {Cross}_\phi(e_{main}, e_{aux}), e_{out} \in \mathbb{R}^{l_{out} \times h_{out} \times w_{out}}
\end{equation}

Subsequently, we can determine the class assignment based on the feature embeddings using a distance metric. In a typical setting, metric-based methods can use the nearest-neighbor approach to determine the assignment of the class by calculating the distance between the query image and the class representatives on the embedding space. Our proposed method can be applied to any metric-based model, so the definition of class representatives is determined by the base model. In ProtoNet, for a $K$-way-$N$-shot problem, the class representative is defined as the mean of the feature embeddings of that class, as defined in Equation~\eqref{eq:mean_proto}. Assume that the prototype embedding of the $k$th class in the support set $S$ is $e_k^p$. Then, for the $j$th image in the query set $Q$ with the feature embedding $e_j^q$, the distance over the class prototypes $e_k^p$ is defined in Equation~\eqref{eq:class_determination}.

\begin{equation}
\label{eq:mean_proto}
e_k^p = \frac{1}{N}\sum_{n=1}^{N} e_{\{k, n\}}^p , k \in \{1,2,\ldots,K\}
\end{equation}

\begin{equation}
\label{eq:class_determination}
d_{\{k,j\}} = \mathcal{F}_\omega(e_k^p, e_j^q), k \in \{1,2,\ldots,K\}, j \in \{1,2,\ldots,M\},
\end{equation}

where $\mathcal{F}_\omega$ is the distance function. In ProtoNet, $\mathcal{F}_\omega$ is the Euclidean distance. Then, for the $j$th query image $q_j$ with embedding $e_j^q$, the final distribution over classes should be:
\begin{equation}
\label{eq:softmax}
p_\theta(y=i|q_j) = \frac{exp(-d_{\{k,j\}})}{\sum_{k=1}^{K} exp(-d_{\{k,j\}})}, i \in \{1,2,\ldots, K\}, j \in \{1,2,\ldots,M\},
\end{equation}
where $M$ is the number of query images in each class defined in a $K$-way-$N$-shot few-shot classification problem.

The visual feature map is available for the auxiliary task once the input image $x$ passes through the embedding network $E_\theta$. Subsequently, we reshape the generated visual feature map into a patch sequence and use a fully-connected layer $FC_\sigma$ to project the visual embedding vector into the space of the same dimension length as the soft labels. Finally, we use a sum operation across the dimension $h \times w$ to make the output have the same dimension as the soft label. The entire process is defined in Equation~\eqref{eq:aux_process1} and Equation~\eqref{eq:aux_process2}.

\begin{equation}
\label{eq:aux_process1}
e_{aux} = FC_\sigma(\mathcal{E}_\theta(x)), x \in \mathbb{R}^{ch_{in} \times h_{in} \times w_{in}}, e_{aux} \in \mathbb{R}^{l_{out} \times (h_{out} \times w_{out})}
\end{equation}
\begin{equation}
\label{eq:aux_process2}
\hat{y}_{aux} = \mbox{softmax}(\mbox{sum}(e_{aux})/\tau), \hat{y}_{aux} \in \mathbb{R}^{l_{out}},
\end{equation}

where $l_{out}$ is the vector length of a soft label, $ch_{in}$ is the number of channels in the input image and $\tau$ is the temperature parameter. In particular, $e_{aux}$ defined in Equation~\eqref{eq:aux_process1} acts as an input to the proposed CAM, and semantic soft labels are the word-embedding vectors of the labels leveraging the pre-trained Glove model~\citep{pennington2014glove}. The semantic soft labels are the targets of the auxiliary task, and the purpose is to place an additional constraint on the main task to tackle the visual-semantic mismatch problem.

\begin{figure}[htp]
\begin{center}
  \includegraphics[width=\textwidth]{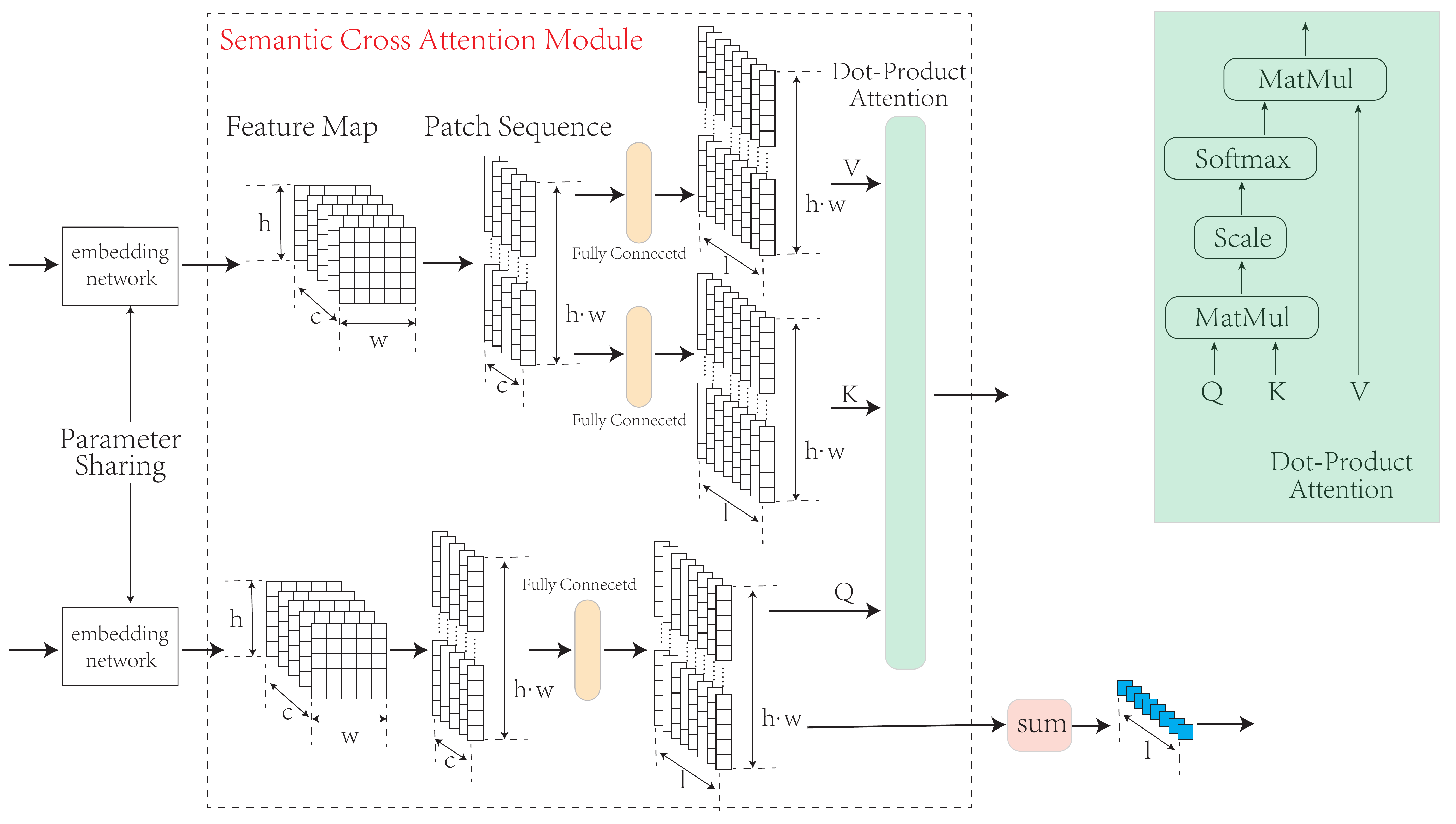}
  \caption{The architecture of semantic cross attention module.}
  \label{fig:cross_attention_module}
\end{center}  
\end{figure}

The model can incorporate semantic and visual features into the network by sharing the embedding network of the two tasks, and we also introduce a CAM to inject semantic features into visual features, as shown in Figure~\ref{fig:cross_attention_module}. As defined in Equation~\eqref{eq:cross_attention}, the CAM takes two inputs, namely, the feature map of the main task $e_{main}=\mathcal{E}_\theta(x)$ and the feature map of the auxiliary task $e_{aux}$. For the feature map of the main task $e_{main}$, it is first reshaped into a patch sequence, then the patch sequence is projected into a value sequence $p_{value}$ and a key sequence $p_{key}$ by two fully-connected layers separately. Then the value sequence, the key sequence, and the query sequence are fed into the Dot-Product Attention module, which is defined by Equation~\eqref{eq:dot_product_attention}, to obtain the final output embedding.

\begin{equation}
\label{eq:dot_product_attention}
e_{out} = \mbox{MatMul}(\mbox{softmax}(\mbox{MatMul}(p_{query}, p_{key}) \cdot scale), p_{value})
\end{equation}

Finally, we use KL-divergence as the loss function of the auxiliary task to calculate the difference between the soft labels and the outputs of the auxiliary task, and we use cross-entropy as the loss function of the few-shot classification task. Therefore, the proposed model has two tasks and the total loss $\mathcal{L}$ of the proposed method is defined in Equation~\eqref{eq:final_loss}, in which $\mathcal{L}_{aux}$ is the loss of the auxiliary task and $\mathcal{L}_{cls}$ is the loss of the few-shot classification task, and $\lambda$ is a hyperparameter that controls the weight of the auxiliary loss.

\begin{equation}
\label{eq:final_loss}
\mathcal{L} = (1 - \lambda)\mathcal{L}_{cls} + \lambda\mathcal{L}_{aux},    
\end{equation}

\section{Experiments}
\label{experiments}
\subsection{Experimental settings \& Results}
We follow the evaluation protocol of the study~\citep{chen2019closerfewshot} by using CUB~\citep{WelinderEtal2010}, mini-ImageNet~\citep{vinyals2016matching}, and tiered-ImageNet~\citep{ren2018meta} to evaluate the proposed method. The mini-ImageNet dataset comprises 64 classes for training, 16 classes for validation, and 20 classes for testing, and each class has 600 images. Meanwhile, the CUB dataset comprises 11,788 images, among which 100 classes were for training, 50 classes for validation, and 50 classes for testing. The tiered-IamgeNet dataset uses 351 classes from 20 categories as the training set, 97 classes from 6 different categories as the validation set, and 160 classes from 8 different categories as the testing set.

Note that the embedding network plays a key role in the metric-based methods for FSL. Consequently, we fix the size of the embedding network using the Conv-4-128 architecture in the experiments for a fair comparison. The Conv-4-128 network comprises four convolution layers that have 64, 64, 128, and 128 filters, respectively. In particular, for the auxiliary task, we add an extra fully-connected layer right after the embedding network to align the visual features and the corresponding text features when they are used to calculate the auxiliary loss, as shown in Figure~\ref{fig:overal_architecture}.

To transform the text of the label into soft labels of the auxiliary task, we use Glove~\citep{pennington2014glove} pretrained on the Common Crawl corpus~\citep{commoncrawl}, and the word embedding vector is of size 300. For the label text that comprises more than one word, we use the average of the embedding vectors for the words involved in the label text as the soft label. All images are randomly resized and cropped to the size of $84 \times 84$, and color jitter and random flip are used as augmentation methods. The $\lambda$ in Equation~\ref{eq:final_loss} is 0.1 in the experiments. We apply the proposed method to two metric-based methods, including ProtoNet~\citep{snell2017prototypical} and ProxyNet~\citep{xiao2020proxy}. The Optimizers are AdamW~\citep{Loshchilov2019DecoupledWD} with an initial learning rate of 0.001 for ProtoNet and SGD with an initial learning rate of 0.1 for ProxyNet. The reduce-lr-on-plateau~\citep{pytorchdocument} method is applied to adjust the learning rate of the optimizer during the training of the two optimizers. We set the number of epochs to 2000, each of which comprises 100 episodes, and the model can be trained in ten hours with an NVIDIA-1080 Ti GPU. The experimental results on the 5-way-1-shot and 5-way-5-shot tasks are listed in Table~\ref{table:FSL_experimental_results_1shot} and Table~\ref{table:FSL_experimental_results_5shot}, in which the three metric-based methods with plus symbols are those enhanced by our proposed method. The values in Table~\ref{table:FSL_experimental_results_1shot} and Table~\ref{table:FSL_experimental_results_5shot} are the accuracy and the 95\% confidence interval, and the performance values for ProtoNet are directly obtained from \citep{chen2019closerfewshot}. 

Although Self-distill~\citep{tian2020rethinking}, DSN-MR~\citep{simon2020adaptive}, and BOIL~\citep{Oh2021BOILTR} are three state-of-the-art methods, Self-distill and DSN-MR do not report their experimental results in the CUB dataset. In Self-distill~\citep{tian2020rethinking}, the knowledge distillation method is defined as a series of generations based on the timeline, which means that the model at time $T$ can learn knowledge from the model at time $T-1$. BOIL is a gradient-based approach that relies on representation change, which is a crucial component in gradient-based methods.

As shown in Table~\ref{table:FSL_experimental_results_1shot} and Table~\ref{table:FSL_experimental_results_5shot}, the experimental results indicate that the two metric-based methods can benefit from our proposed method and produce significant improvement. Moreover, using our proposed method can allow these methods to achieve competitive performance compared to state-of-the-art methods. The source code for our model is publicly accessible, and detailed hyperparameters and settings are available in the source code.

We focus on how to train a better embedding network as it is a crucial component for metric-based methods. Central to our proposed method is to consider the semantic information obtained from the text labels as an auxiliary task to incorporate semantic information into the training of the embedding network. An ablation study is conducted to analyze the importance of these components in our proposed method.

\begin{table*}[t]
\caption{Experimental results on CUB, mini-ImageNet, and tiered-ImageNet datasets for 5-way-1-shot tasks. $+$ means the model trained with the proposed method using ConvNet4-128 backbone, {\ddag} means our re-implementation of ProtoNet and ProxyNet with ConvNet4-128 backbone.}
\centering
\begin{tabular}{ c c c c c c}
\hline
\label{table:FSL_experimental_results_1shot}
Method & CUB & mini-ImageNet & tiered-ImageNet\\
\hline
 \texttt{ProtoNet} & $50.46\pm 0.88$ & $44.42 \pm 0.84$  & $-$  \\
 \texttt{ProxyNet} & $67.52\pm 0.97$ & $52.95 \pm 0.76$  & $-$ \\
 \texttt{Self-distill} & $-$ & $55.88 \pm 0.59$ & $56.76 \pm 0.68$ & \\
 \texttt{DSN-MR} & $-$ & $55.88 \pm 0.90$ & $-$ \\
 \texttt{BOIL} & $61.60 \pm 0.57$ & $49.61 \pm 0.16$ & $48.58 \pm 0.27$\\
 \hline
 \texttt{ProtoNet$^{\ddag}$} & $62.67 \pm 0.93$ & $ 50.21 \pm 0.82 $ & $43.62 \pm 0.82 $ \\
 \texttt{ProxyNet$^{\ddag}$} & $72.36 \pm 0.84$ & $54.40 \pm 0.81$ & $55.05 \pm 0.93$ \\
 \hline
 \texttt{ProtoNet$^+$} & $ 74.28 \pm 0.87 $ & $ 54.34 \pm 0.81 $ & $ 56.25 \pm 0.97 $ \\
 \texttt{ProxyNet$^+$} & $ \mathbf{76.66 \pm 0.87} $ & $ \mathbf{57.45 \pm 0.86} $ & $ \mathbf{59.32 \pm 0.90} $ \\
\hline
\end{tabular}
\end{table*}

\begin{table*}[t]
\caption{Experimental results on CUB, mini-ImageNet, and tiered-ImageNet datasets for 5-way-5-shot tasks. $+$ means the model trained with the proposed method using ConvNet4-128 backbone, {\ddag} means our re-implementation of ProtoNet and ProxyNet with ConvNet4-128 backbone.}
\centering
\begin{tabular}{ c c c c c c}
\hline
\label{table:FSL_experimental_results_5shot}
  Method & CUB & mini-ImageNet & tiered-ImageNet\\
\hline
 \texttt{ProtoNet} & $76.39 \pm 0.64$ & $64.24\pm 0.72$ & $-$ \\
 \texttt{ProxyNet} & $82.85 \pm 0.60$ & $70.35 \pm 0.63$ & $-$\\
 \texttt{Self-distill} & $-$ & $71.65 \pm 0.51$ &  $73.21 \pm 0.54$\\
 \texttt{DSN-MR} & $-$ & $ 70.50 \pm 0.68$ & $-$\\
 \texttt{BOIL} & $75.96 \pm 0.17$ & $ 66.45 \pm0.37$ & $69.37 \pm 0.12$\\
 \hline
 \texttt{ProtoNet$^{\ddag}$} & $79.39 \pm 0.65$ & $ 69.64 \pm 0.65 $ & $72.18 \pm 0.76 $\\
 \texttt{ProxyNet$^{\ddag}$} & $86.15 \pm 0.51$ &  $72.36 \pm 0.67$ & $73.77 \pm 0.76$\\
 \hline
 \texttt{ProtoNet$^+$} & $ 86.39 \pm 0.49 $ & $ 73.12 \pm  0.63 $ & $ 76.89 \pm 0.72 $\\
 \texttt{ProxyNet$^+$} & $ \mathbf{88.48 \pm 0.46} $ & $\mathbf{73.54 \pm 0.62}$ & $\mathbf{77.83 \pm 0.69}$ \\
\hline
\end{tabular}
\end{table*}

\begin{figure}[htp]
\begin{center}
  \includegraphics[width=0.8\columnwidth]{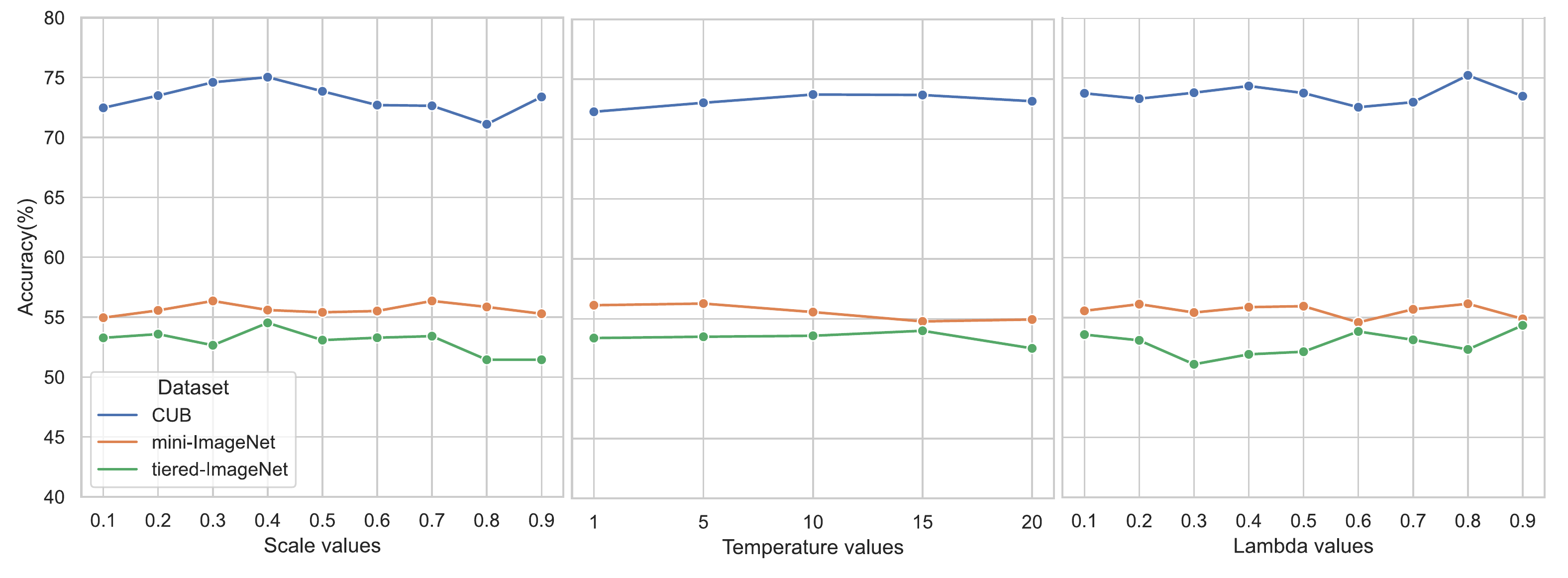}
  \caption{Experimental results with different values of $\lambda$ values, different values of temperature $\tau$ and different values of $scale$.}
  \label{fig:results_on_different_lambda}
\end{center}
\end{figure}

\subsection{Sensitivity Analysis}
To further investigate the impact of the proposed semantic feature alignment on the performance of the model, we perform a sensitivity analysis by changing the value of $\lambda$ in Equation~\ref{eq:final_loss} with ProtoNet as the base model. The experiments are carried out on CUB, mini-ImageNet, and tired-ImageNet datasets for the 5-way-1-shot problem, and the experimental results are shown in Figure~\ref{fig:results_on_different_lambda}. It can be seen from the results that our proposed method produces a relatively stable performance under different values of $\lambda$ in the range from 0.1 to 0.9. Similarly, we also perform sensitivity analysis on the other two hyperparameters, namely the temperature $\tau$ defined in Equation~\ref{eq:aux_process2} and the $scale$ value defined in Equation~\ref{eq:dot_product_attention}, and the experimental results are also shown in Figure~\ref{fig:results_on_different_lambda}. Note that the accuracy scores listed in Figure~\ref{fig:results_on_different_lambda} are obtained using the validation set. Overall, the experimental results show that the proposed method is robust to different choices of these hyper-parameters, especially for $\tau$, in the range reported in Figure~\ref{fig:results_on_different_lambda}.

\subsection{Ablation Study}
We conduct an ablation study to assess the effectiveness of the components involved in our proposed method. First, we conduct experiments to verify whether the main few-shot classification task in our model can benefit from the proposed CAM and the auxiliary task. We compare our model with the multi-task learning model without CAM. Furthermore, the squeeze-and-excitation network~\citep{hu2018squeeze} is a simple but useful module on channel attention, so we also develop a model that replaces our proposed CAM module with a squeeze-excitation module to re-weight the importance of the channel, but without using the inputs of aligned semantic features that are from the auxiliary task.

\begin{figure}[htp]
\begin{center}
  \includegraphics[width=0.9\columnwidth]{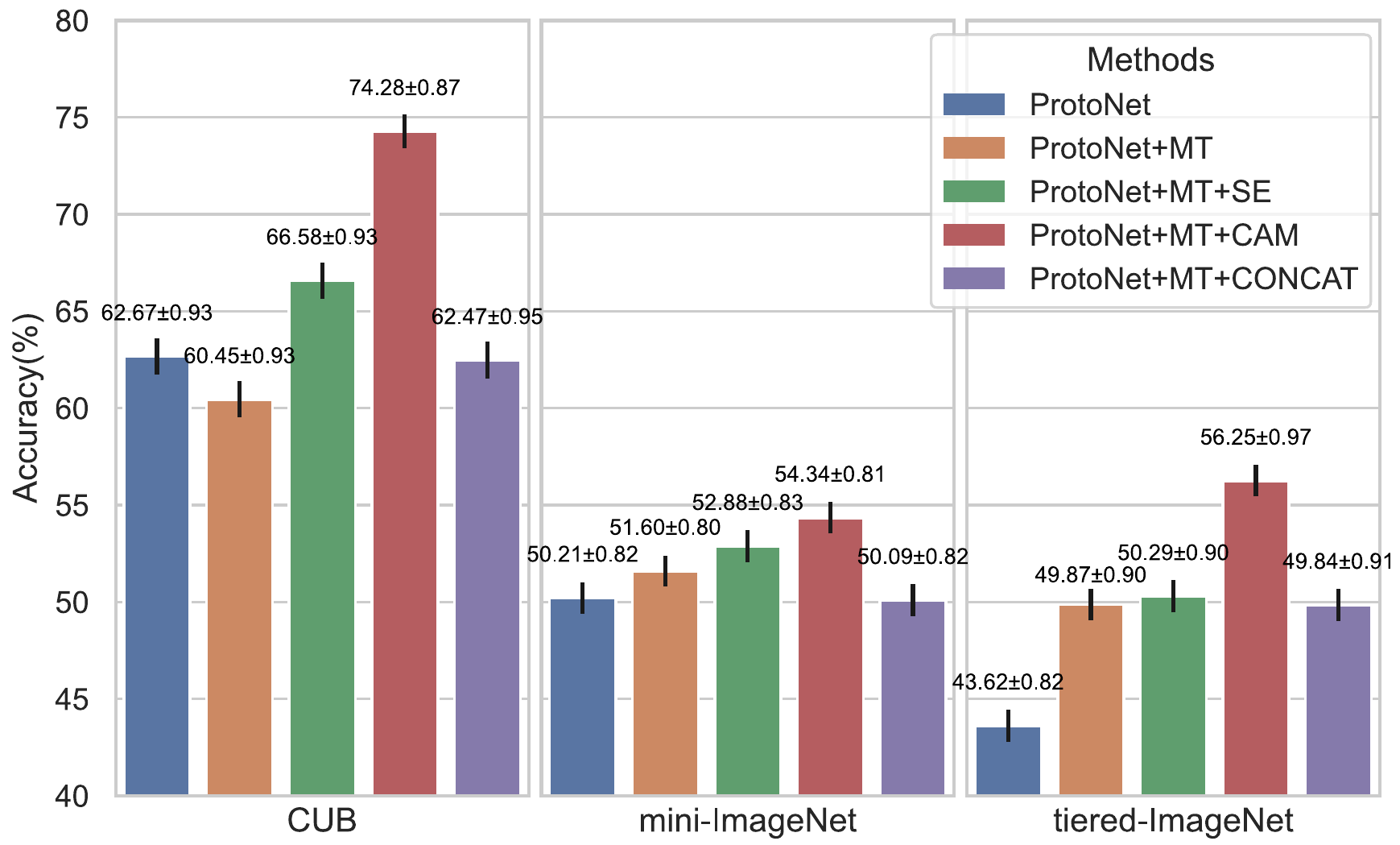}
  \caption{The ablation study results for 5-way-1-shot tasks. ProtoNet here is our re-implemented version; ProtoNet+MT means the model trained only with the multi-task architecture without using the proposed CAM component; ProtoNet+MT+CAM means the model trained with the multi-task architecture and the proposed CAM component. ProtoNet+MT+CONCAT means the model trained with the multi-task architecture and a concatenation of the feature maps from the main task and the auxiliary task is used for the subsequent classification.}
  \label{fig:ablation_results_on_1shot}
\end{center}  
\end{figure}

Similarly, we still use ProtoNet as the base model, and the experiments are carried out on CUB, mini-ImageNet, and tiered-ImageNet datasets for the 5-way-1-shot problem. The experimental results are presented in Figure~\ref{fig:ablation_results_on_1shot}. The experimental results point out that only using the multi-task learning architecture and semantic features can significantly improve the performance. Thus, this can empirically show the importance of semantic features in our proposed method. Moreover, both the squeeze-excitation module and the proposed CAM can improve performance. The proposed CAM module can outperform the squeeze-excitation module in all three data sets. We also list a model implemented in the proposed multi-task manner, but the semantic features generated in the auxiliary task are concatenated with the visual features in the main task. The results show that this simple concatenation method of merging visual features and semantic features only leads to performance improvement on the tiered-ImageNet dataset, demonstrating that our proposed CAM module is adequate for combining visual and semantic features.

\subsection{Embedding Network}
We also explore the impact of the embedding network using ResNet12~\citep{he2016deep} to replace the Conv-4-128 network. Many studies have shown that a deeper embedding network may improve performance. We conduct experiments on CUB, mini-ImageNet, and tiered-ImageNet datasets for 5-way-1-shot and 5-way-5-shot problems using ProtoNet and ProxyNet with ResNet-12 as their embedding networks. The experimental results are listed in Table~\ref{table:FSL_different_backbone_results_1shot} and Table~\ref{table:FSL_different_backbone_results_5shot}, indicating that the use of ResNet-12 can significantly improve the performance of ProtoNet and ProxyNet. Thus, the proposed method can also benefit from a deeper embedding network.

\begin{table*}[t]
\caption{Experimental results for 5-way-1-shot tasks with different embedding networks. ${+}$ means the proposed model trained with the ConvNet4-128 backbone, and ${\dag}$ means the proposed model trained with the ResNet12 backbone.}
\begin{center}
\begin{tabular}{ c c c c}
\hline
\label{table:FSL_different_backbone_results_1shot}
  Method & CUB & mini-ImageNet & tiered-ImageNet\\
 \hline
 \texttt{ProtoNet$^+$} & $ 74.28 \pm 0.87 $  & $ 54.34 \pm 0.81 $ & $ 56.25 \pm 0.97 $\\
 \texttt{ProxyNet$^+$} & $ 76.66 \pm 0.87 $  & $ 57.45 \pm 0.86 $ & $ 59.32 \pm 0.90 $\\
 \hline

 \texttt{ProtoNet$^{\dag}$} & $ 79.81 \pm 0.84 $ & $ 57.89 \pm 0.85 $ & $ 60.40 \pm 0.98 $ \\
 \texttt{ProxyNet$^{\dag}$} & $ \mathbf{80.20 \pm 0.81} $ & $ \mathbf{60.58 \pm 0.85} $  & $ \mathbf{63.84 \pm 1.02} $ \\
  
\hline
\end{tabular}
\end{center}
\end{table*}

\begin{table*}[t]
\caption{Experimental results for 5-way-5-shot tasks with different embedding networks. ${+}$ means the proposed model trained with the ConvNet4-128 backbone, and ${\dag}$ means the proposed model trained with the ResNet12 backbone.}
\begin{center}
\begin{tabular}{ c c c c }
\hline
\label{table:FSL_different_backbone_results_5shot}
 Method & CUB & mini-ImageNet & tiered-ImageNet\\
 \hline
 \texttt{ProtoNet$^+$} & $ 86.39 \pm 0.49 $ & $ 73.12 \pm  0.63 $ & $ 76.89 \pm 0.72 $\\
 \texttt{ProxyNet$^+$}  & $ 88.48 \pm 0.46 $ & $ 73.54 \pm 0.62$  & $ 77.83 \pm 0.69$ \\
 \hline

 \texttt{ProtoNet$^{\dag}$} & $ 89.10 \pm 0.42 $ & $ 75.56 \pm 0.63 $  & $ \mathbf{83.60 \pm 0.67} $\\
 \texttt{ProxyNet$^{\dag}$} & $ \mathbf{90.80 \pm 0.39} $  & $ \mathbf{76.53 \pm 0.58} $ & $ 83.15 \pm 0.69 $\\
  
\hline
\end{tabular}
\end{center}
\end{table*}

\subsection{Auxiliary task with different loss functions}
In our proposed method, we use KL loss as the loss of the semantic task. Other loss functions such as mean squared loss (MSE) can also be used in the semantic task. In Table~\ref{table:FSL_different_loss_results_1shot} and Table~\ref{table:FSL_different_loss_results_5shot}, we compare the performance of using the KL loss and the MSE loss as an auxiliary loss in the semantic task. We set $\lambda$ in Equation~\ref{eq:final_loss} to 0.1 in the experiments. The experimental results show that MSE loss can also work well with the proposed method, and the performance of KL loss and MSE loss are close, but overall, MSE loss does not show any superiority compared to KL loss.

\subsection{Auxiliary task with different pre-trained language models}
In this study, we propose using a pre-trained language model to generate the soft labels used in the auxiliary task. To evaluate the impact of different pre-trained language models, we conduct experiments with three classic pre-trained language models, including Glove~\citep{pennington2014glove}, FastText~\citep{bojanowski2017enriching} and Bert~\citep{devlin2018bert} for the 5-way-1shot problem. The experimental results are listed in Table~\ref{table:different_pretrained_models}. The experimental results show that Glove generally performs better than the other two.

\begin{table*}[t]
\caption{Experimental results for 5-way-1-shot tasks with different auxiliary losses. ${+}$ means the proposed model trained with KL loss as the auxiliary loss, ${\ast}$ means the proposed model trained with MSE loss as the auxiliary loss.}

\begin{center}
\begin{tabular}{ c c c c}
\hline
\label{table:FSL_different_loss_results_1shot}
 Method & CUB & mini-ImageNet & tiered-ImageNet\\
 \hline
 \texttt{ProtoNet$^+$} & $ 74.28 \pm 0.87 $ & $ 54.34 \pm 0.81 $ & $ 56.25 \pm 0.97 $ \\
 \texttt{ProxyNet$^+$} & $ \mathbf{76.66 \pm 0.87} $ & $ \mathbf{57.45 \pm 0.86} $ &  $ \mathbf{59.32 \pm 0.90} $ \\
 \hline

 \texttt{ProtoNet$^{\ast}$} & $ 74.82 \pm 0.87 $ & $ 55.05 \pm 0.86 $ & $ 54.89 \pm 0.94 $\\
 \texttt{ProxyNet$^{\ast}$} & $ 75.98 \pm 0.88 $ & $ 55.80 \pm 0.86 $ & $ 58.59 \pm 0.96 $\\
\hline
\end{tabular}
\end{center}
\end{table*}

\begin{table*}[t]
\caption{Experimental results for 5-way-5-shot tasks with different auxiliary losses. ${+}$ means the proposed model trained with KL loss as an auxiliary loss, ${\ast}$ means the proposed model trained with MSE loss as an auxiliary loss.}

\begin{center}
\begin{tabular}{ c c c c}
\hline
\label{table:FSL_different_loss_results_5shot}
  Method & CUB & mini-ImageNet & tiered-ImageNet\\
 \hline
 \texttt{ProtoNet$^+$}  & $ 86.39 \pm 0.49 $ & $ 73.12 \pm  0.63 $ & $ 76.89 \pm 0.72 $\\
 \texttt{ProxyNet$^+$}  & $ \mathbf{88.48 \pm 0.46} $ & $ \mathbf{73.54 \pm 0.62} $ & $ \mathbf{77.83 \pm 0.69} $ \\
 \hline

 \texttt{ProtoNet$^{\ast}$} & $ 86.35 \pm 0.49 $ & $ 72.52 \pm 0.63 $  & $ 76.41 \pm 0.73 $\\
 \texttt{ProxyNet$^{\ast}$} & $ 87.59 \pm 0.49 $ & $ 73.12 \pm 0.66 $  & $ 77.59 \pm 0.72 $\\
\hline
\end{tabular}
\end{center}
\end{table*}

\begin{table}[t]
\caption{Experimental results for 5-way-1-shot tasks with different word embedding models. $\dag$ means using ProtoNet as the base model, while $\ddag$ means that the base model is ProxyNet.}
\begin{center}
\begin{tabular}{ c c c c}
\hline
\label{table:different_pretrained_models}
  Method & CUB & mini-ImageNet & tiered-ImageNet\\
 \hline
 \texttt{Glove$^\dag$} & $ 74.28 \pm 0.87 $  & $ 54.34 \pm 0.81 $ & $ 56.25 \pm 0.97 $\\
 \texttt{FastText$^\dag$} & $ 74.82 \pm 0.82 $  & $ 52.79 \pm 0.83 $ & $ 52.20 \pm 0.91 $\\
 \texttt{Bert$^\dag$} & $ 74.39 \pm 0.86 $  & $ 53.45 \pm 0.86 $ & $ 53.46 \pm 0.94 $\\
 \hline
 \texttt{Glove$^\ddag$} & $ \mathbf{76.66 \pm 0.87} $  & $ \mathbf{57.45 \pm 0.86} $ & $ \mathbf{59.32 \pm 0.90} $\\
 \texttt{FastText$^\ddag$} & $ 74.64 \pm 0.85 $  & $ 55.56 \pm 0.84 $ & $ 55.81 \pm 0.88 $\\
 \texttt{Bert$^\ddag$} & $ 75.63 \pm 0.90 $  & $ 56.35 \pm 0.86 $ & $ 54.82 \pm 0.92 $\\
  
\hline
\end{tabular}
\end{center}
\end{table}

\section{Conclusions}
This study presents the visual-semantic mismatch problem that is present in the FSL setting and proposes to incorporate semantic features into the model to help alleviate the problem of only limited images available for each episode. In our proposed multi-task learning model, we align the visual features with their semantic meanings to enforce the model to consider the semantic meaning of text labels and visual features simultaneously, giving a base to train an embedding network that can quickly adapt to novel classes with limited training samples. Besides, we propose a CAM module to further refine the visual embeddings by using the aligned features of the auxiliary task. We apply the proposed method to two classical metric-based methods. The experimental results indicate that these methods can benefit from our proposed method to achieve substantial improvement and competitive performance compared to state-of-the-art models. Our proposed method can also be applied to other metric-based methods without changing the pipeline of the original methods.

\acks{This work was supported in part by Ministry of Science and Technology, Taiwan, under Grant no. MOST 109-2628-E-009-009-MY3 and 111-2221-E-A49-083-MY3. We are grateful to the National Center for High-performance Computing for computer time and facilities.
}


\bibliography{acml22}

\end{document}